\documentclass[runningheads]{llncs}
\usepackage[T1]{fontenc}
\usepackage{graphicx}

\usepackage{subfigure} 

\usepackage[hidelinks]{hyperref}

\usepackage{color}

\begin{document}

\title{\texttt{LARS}: Light Augmented Reality System for Swarm}

\titlerunning{\texttt{LARS}}
\author{Mohsen Raoufi\inst{1,2} \and 
Pawel Romanczuk \inst{1,2} \and 
Heiko Hamann\inst{1,3} 
}
\authorrunning{M. Raoufi et al.}
\institute{Science of Intelligence, Research Cluster of Excellence, Berlin, Germany \email{mohsenraoufi@icloud.com}
\and 
Department of Biology, Humboldt University of Berlin, Berlin, Germany
\and
Department of Computer and Information Science, University of Konstanz, Konstanz, Germany\\
}
\index{Raoufi, Mohsen}
\index{Romanczuk, Pawel}
\index{Hamann, Heiko}
\maketitle              
\setcounter{footnote}{0}

Extended reality (XR) technology has found its applications in various systems, including multi-robot systems~\cite{makhataeva_augmented_2020}. 
Augmented Reality and Mixed Reality tools are becoming increasingly influential in educational technology and robotics. They enrich learning experiences and expand research methodology. Multi-robot systems, as a complex collective system, have great potential for educational purposes, showcasing collective behaviors. In such systems, XR tools set up a dynamic virtual environment observable by humans as well as a medium to interact with multi-robot systems. Making non-tangible and complex concepts in multi-robot systems easier to comprehend by human users will help in understanding, analyzing, and conveying how these systems do what they do. 

We present the Light Augmented Reality System (\texttt{LARS}) as an open-source \href{https://github.com/mohsen-raoufi/LARS}{online}\footnote[1]{\url{https://github.com/mohsen-raoufi/LARS}} and cost-effective tool. 
\texttt{LARS} leverages light-projected visual scenes for indirect robot-robot and human-robot interaction through the real environment. It operates in real-time and is compatible with a range of robotic platforms, from miniature to middle-sized robots.
\texttt{LARS} can support researchers in conducting experiments with increased freedom, reliability, and reproducibility. This XR tool makes it possible to enrich the environment with full control by adding complex and dynamic objects while keeping the properties of robots as realistic as they are. 
The system promotes stigmergy as a natural method of indirect communication between robots. However, it also keeps the possibility of directly transmitting messages to robots using central information.
Furthermore, such interactive systems set the scene to investigate the principles across various disciplines, e.g., biology and social sciences, as we used in our previous studies~\cite{raoufi2023individuality}. 

\begin{figure*}[!t]
\centering
    \subfigure[]{\includegraphics[height=1.2 in]{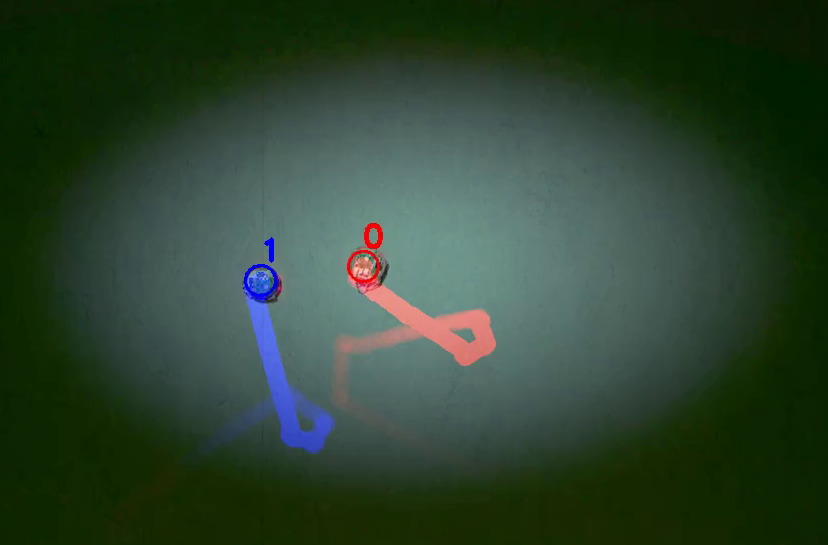}}
    \subfigure[]{\includegraphics[height=1.2 in]{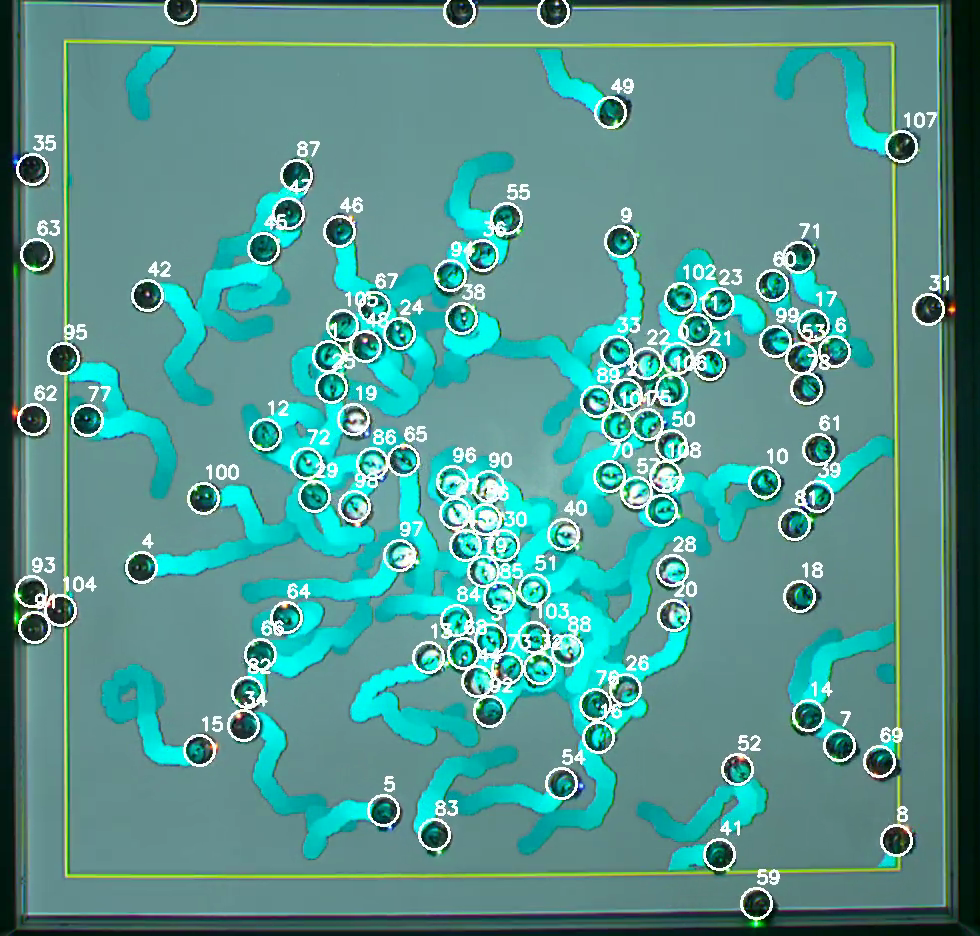}}
    \subfigure[]{\includegraphics[height=1.2 in]{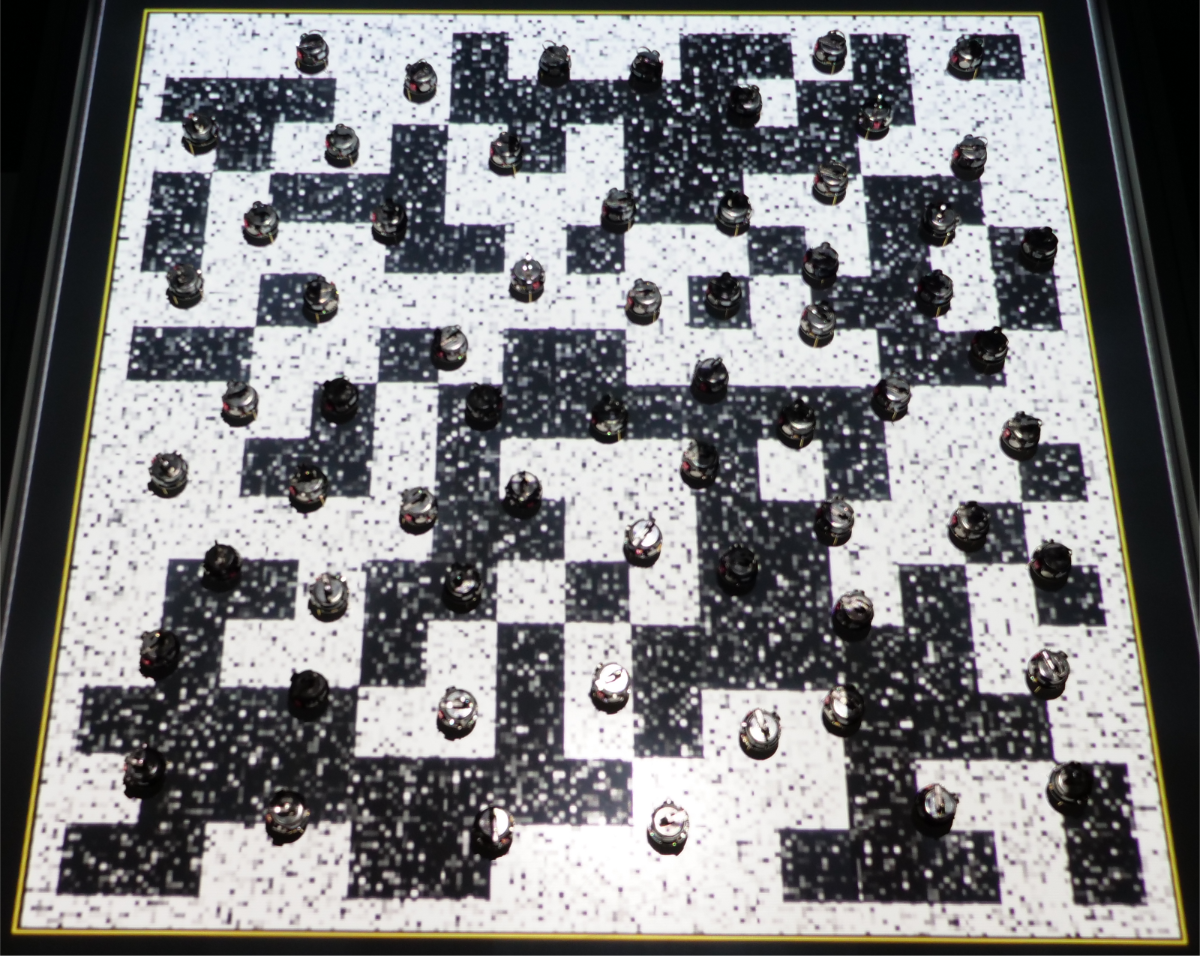}}
    
\caption{Example scenarios with (a) two Thymio robots with different ring colors, (b)~109 Kilobots cleaning up the environment, and (c) 63 Kilobots making a collective decision on a tiled environment with dynamic noise. }
\label{fig:example_scenarios}
\end{figure*}

Key features of \texttt{LARS} include light-based interaction, marker-free cross-\linebreak platform tracking, real-time performance, scalability, and ease of setup. The system uses light as a medium for (indirect) communication. Indirect communication via the environment, Stigmergy, is the key to making \texttt{LARS} independent of a specific platform, compared to the previous robot-specific tools~\cite{reina_ark_2017}. Light is easily observable by both humans and robots and allows for human-robot interaction in the \textit{real} world. This is in contrast to other related works that simulate a virtual environment for the interaction of the system~\cite{reina_ark_2017}. Marker-free tracking enhances the versatility of the system and reduces the time and effort needed to integrate new platforms. \texttt{LARS} operates in real-time with approximately 38 frames per second on a standard desktop PC. 
We implemented the software in 
\texttt{C++}
language and used the OpenCV library due to their real-time performance and being free and open-source. 
At the core of the detection and tracking code we employed and optimized the tracker developed for \texttt{ARK}~\cite{reina_ark_2017}, which was originally developed specifically for Kilobots~\cite{rubenstein2012Kilobot}. Given the prevalent circular shape of swarm robot platforms, \texttt{LARS} harnesses this regularity and detects various robotic platforms using a simple Hough method for circle detection.

\texttt{LARS} is a test-bed system and a tool to conduct multi-robot experiments with; a recording, tracking, and logging system to save experiment data; and a medium through which the reality is extended by augmenting virtual objects. Without altering the limitation of a robotic system, \texttt{LARS} enriches the environment, for example, by adding user-defined noise. The augmented environment is a step toward bridging the gap between simulation and reality. 
We show examples in Fig.~\ref{fig:example_scenarios}, where (a) Thymio robots explore the environment, leaving an elusive pheromone trail on the environment, (b) Kilobots diffuse in the environment to clean it up, and (c) Kilobots make a collective decision in a noisy environment.

\subsubsection*{Acknowledgements.}
Funded by the Deutsche Forschungsgemeinschaft (DFG, German Research Foundation) under Germany’s Excellence Strategy – EXC 2002/1 “Science of Intelligence” – project number 390523135.
%
%
\bibliographystyle{splncs04}
\bibliography{mybibliography}

\begin{thebibliography}{1}
\providecommand{\url}[1]{\texttt{#1}}
\providecommand{\urlprefix}{URL }
\providecommand{\doi}[1]{https://doi.org/#1}

\bibitem{makhataeva_augmented_2020}
Makhataeva, Z., Varol, H.A.: Augmented reality for robotics: A review. Robotics  \textbf{9}(2), ~21 (2020)

\bibitem{raoufi2023individuality}
Raoufi, M., Romanczuk, P., Hamann, H.: Individuality in swarm robots with the case study of kilobots: Noise, bug, or feature? In: ALIFE 2023: Proceedings of the 2023 Artificial Life Conference. MIT Press (2023)

\bibitem{reina_ark_2017}
Reina, A., Cope, A.J., Nikolaidis, E., Marshall, J.A., Sabo, C.: {ARK}: {Augmented} reality for {Kilobots}. IEEE Robotics and Automation letters  \textbf{2}(3) (2017)

\bibitem{rubenstein2012Kilobot}
Rubenstein, M., Ahler, C., Nagpal, R.: {K}ilobot: A low cost scalable robot system for collective behaviors. In: 2012 IEEE International Conference on Robotics and Automation (ICRA) (2012)

\end{thebibliography}
\end{document}